\documentclass[10pt,conference,compsocconf]{IEEEtran}

\usepackage{hyperref}
\usepackage{graphicx}	
\usepackage{amsmath}
\usepackage{amssymb}
\usepackage{caption}
\usepackage{subfigure}
\usepackage{float}
\usepackage{xcolor}

\title{Using Machine Learning for move sequence visualization and generation in climbing}
\author{\textit{Thomas Rimbot, Martin Jaggi, Luis Barba - EPFL}}

\begin{document}
\maketitle
\thispagestyle{plain}
\pagestyle{plain}
\begin{abstract}
    In this work, we investigate the application of Machine Learning techniques to sport climbing. Expanding upon previous projects, we develop a visualization tool for move sequence evaluation on a given boulder. Then, we look into move sequence prediction from simple holds sequence information using three different Transformer models. While the results are not conclusive, they are a first step in this kind of approach and lay the ground for future work.
\end{abstract}
\section{Introduction}
Applying Machine Learning techniques to competitive sport has been an increasing trend in the past few years. We can for example cite the case of car racing or hockey. In this project, we focus on \textit{bouldering}, a form of rock climbing where athletes are tasked with overcoming a small natural or artificial feature (about $4m$ high), requiring both physical strengths and problem-solving skills. \\ Some research has already been done on the topic ($e.g.$ see \cite{ml_methods_route_classif}). Here, we focus on projects realized by EPFL students (see \cite{move_seq_det} and \cite{understanding_bouldering_ML}) which have looked into pose analysis for move sequence detection applied to bouldering. Here, we aim at expanding this research, providing a more tangible visualization pipeline and experimenting with move sequence prediction from holds information.
\section{Move sequence detection}
The starting point of this project was the work done in \cite{move_seq_det}. In this work, the authors developed a pipeline for move sequence detection starting from climbing videos. More details can be found in their paper but we refer in this section the general procedure.
\subsection{Pose estimation}
Using the \texttt{Mediapipe} \cite{mediapipe} library, the video of a person climbing is analyzed, and this person's pose can be extracted. More precisely, for each frame, the visibility and $x-y$ coordinates of the $33$ landmarks are acquired, amounting to a total of $99$ pose information. The landmarks correspond to specific points on the human body, such as the ankle, knee, elbow $etc.$
\subsection{Static extremity detection}
Once the landmarks information are all recorded, they are aggregated into a dataframe and the points corresponding to common extremities are gathered together. For instance, the left hand extremity is composed of the landmarks corresponding to the left wrist, pinky, index and thumb. By checking the distance between two consecutive positions of an extremity at two different frames, it is possible to evaluate if it was static or not, $i.e.$ if it was stable on a hold or in motion. This way, we end up with a cloud of points for all extremities and frames where they are static.
\subsection{Clustering and visualization}
From this cloud of points, we can use clustering algorithms, such as the DBSCAN \cite{dbscan} to regroup them and evaluate the position of the holds. Using \texttt{OpenCV} \cite{opencv}, we can draw squares on a still picture of the boulder, representing the detected holds. Moreover, since we have access to the time information through the frame number, we can find the order in which the holds are used, which, combined with the extremity, is exactly the \textit{Move Sequence}.
\begin{figure}
    \centering
    \includegraphics[width=5cm, height=6cm]{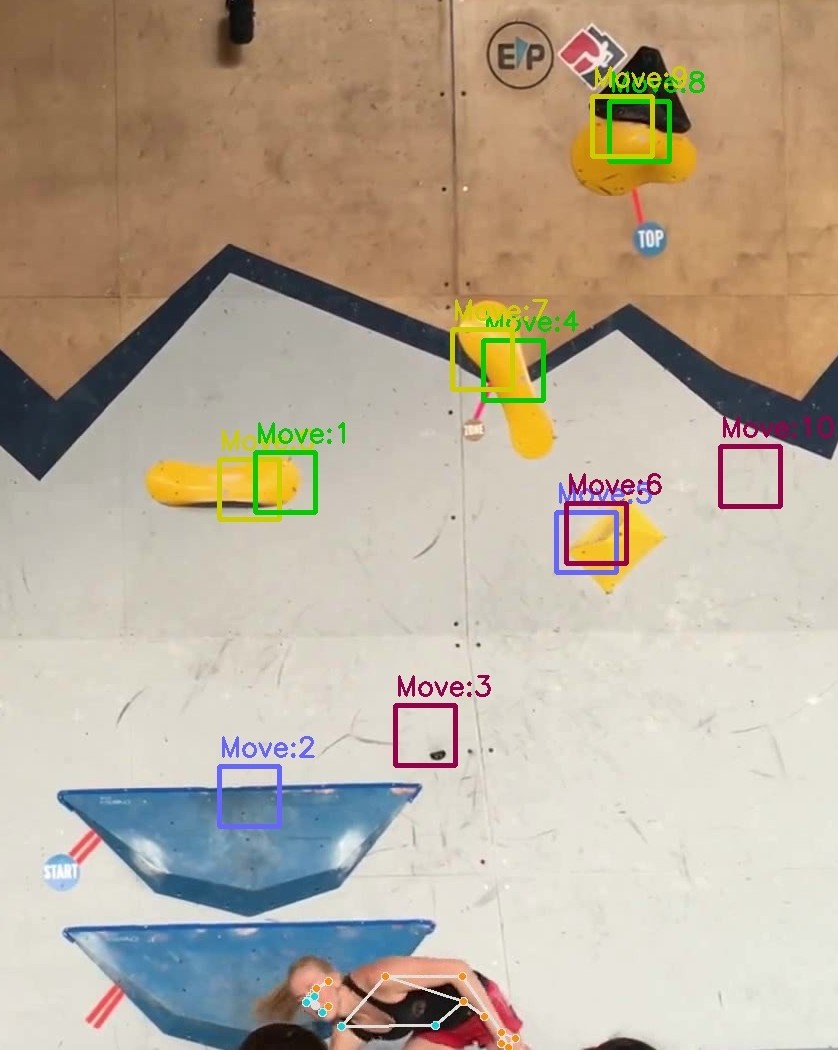}
    \caption{Move sequence detection from \cite{move_seq_det}}
    \label{fig:move_seq}
\end{figure}
\section{Move sequence visualization}
Knowing that working directly with the move sequence was possible, we set out to create a tool which would improve the user experience and allow to visualize it in a more tangible way.
\subsection{Skeleton generation}
\label{ssec:skeleton_generation}
From \cite{move_seq_det}, we can output the move sequence as a sequence of holds coordinates, together with information on the used extremity for each move. This sequence can then be printed on a still image of the boulder, as shown in \autoref{fig:move_seq}. However, it is hard to visualize and imagine an actual human climbing, following this sequence. Therefore, we created a visualization pipeline, with the goal of generating and drawing a humanoid figure using \texttt{Mediapipe}'s features.

More precisely, \texttt{Mediapipe} allows to draw the recorded landmarks, and superimpose them on the video of the person climbing. These points, linked together with edges, form a humanoid skeleton representing a climber. Playing with this tool, we were able to draw the landmarks directly on the image of the boulder instead of the video. 

With this tool in place, the next step was to generate the landmarks information from the move sequence. More precisely, the general form of a move sequence is reported in \autoref{tab:move_seq}. The $x$ and $y$ coordinates represent the position of the hold, the limb is the extremity used (hand or foot, left or right), and the indices indicate the order in which these holds are used. 

\begin{table}[h]
    \centering
    \begin{tabular}{|c|c|c|}
    \hline
        $x_1$ & $y_1$ & $limb_1$ \\
        \hline 
        $x_2$ & $y_2$ & $limb_2$ \\
        \hline 
        $x_3$ & $y_3$ & $limb_3$ \\
        \hline 
        $...$ & $...$ & $...$ \\
        \hline 
        $x_n$ & $y_n$ & $limb_n$ \\
    \hline
    \end{tabular}
    \caption{Move sequence template}
    \label{tab:move_seq}
\end{table}

On the other hand, \texttt{Mediapipe} needs precise information on all 33 landmarks, as explained above, for each frame. So we had to find a way of generating the $99$ features for each frame of the desired video duration (which we usually considered to be about $1500$ frames), from the move sequence alone. This was done in two steps.
\\

\subsubsection{Generation of the landmarks sequence for the extremities}
The move sequence gives us information on the position of the extremities, in a certain order. From this, we can interpolate the positions between consecutive holds to get a continuous motion of the extremities. For instance, if the four extremities are fixed on holds, and we know that the next move in the sequence involves the left hand, we can interpolate the coordinates of the corresponding landmarks between the current hold's and the next one's. This allows to generate a continuous sequence of landmarks coordinates for the left hand, which can then be used to draw the skeleton with the above-described features.
\\
The interpolation speed is weighted by the distance between two consecutive holds of the same extremity and the average number of desired frames for this move, giving a sense of realism to the motion and allowing to tune its duration. 
\\

\subsubsection{Generation of the landmarks sequence for the rest of the body}
Using this procedure, we have a sequence of information for all the extremities landmarks. We are now left to generate the information of the rest of the landmarks, namely the ones corresponding to the rest of the body: the legs, torso, arms and head. In order to do that, we can use the work done in \cite{move_seq_det} for pose detection. In fact, we have access to a lot of video-recorded landmarks sequence of climbers. From this dataset, we can use the extremities landmarks information as features, and try to predict the body landmarks. 

\begin{figure}[h]
     \centering
     \begin{subfigure}
         \centering
         \includegraphics[width=4cm]{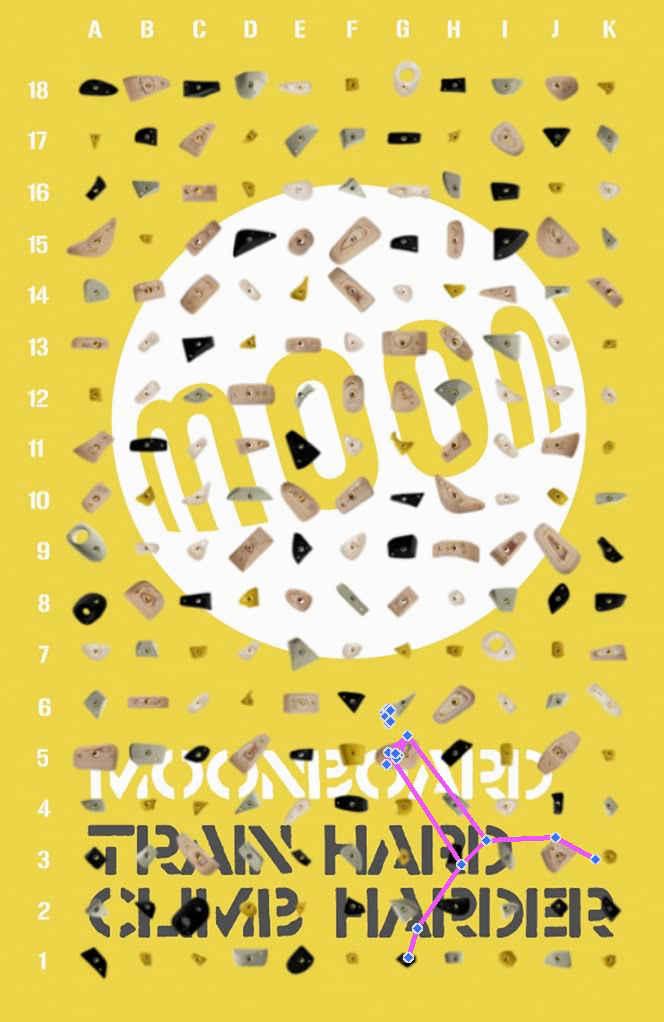}
     \end{subfigure}
     \begin{subfigure}
         \centering
         \includegraphics[width=4cm]{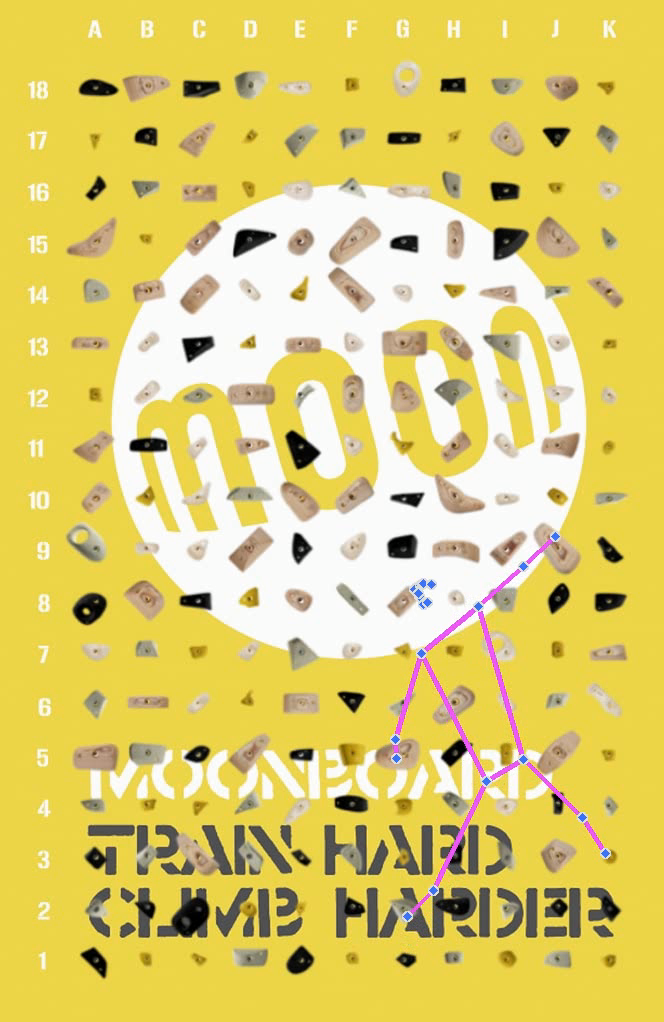}
     \end{subfigure}
     \begin{subfigure}
         \centering
         \includegraphics[width=4cm]{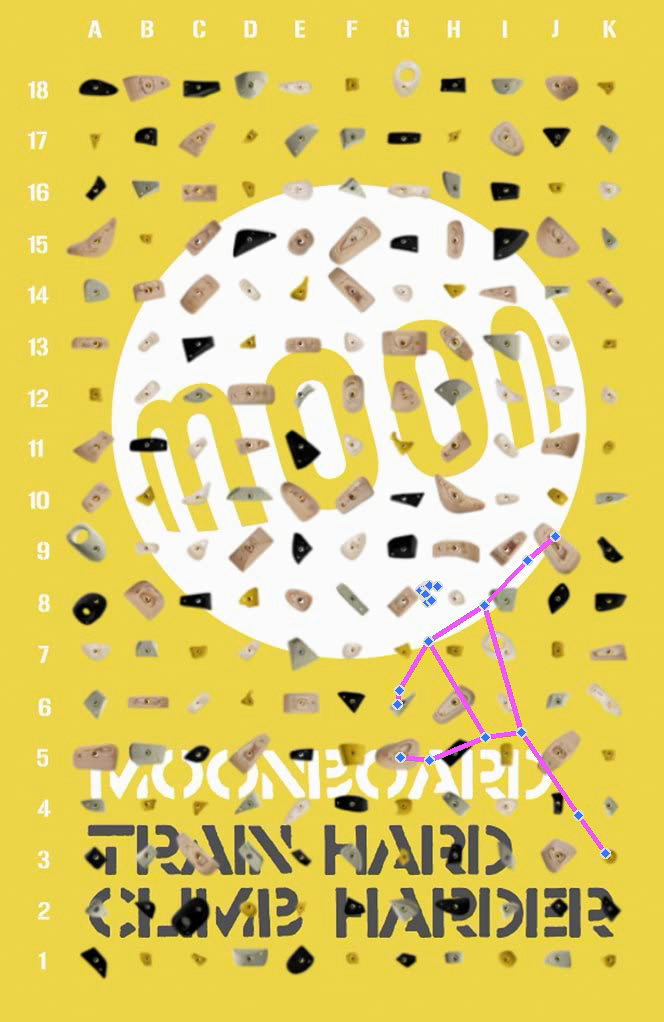}
     \end{subfigure}
     \begin{subfigure}
         \centering
         \includegraphics[width=4cm]{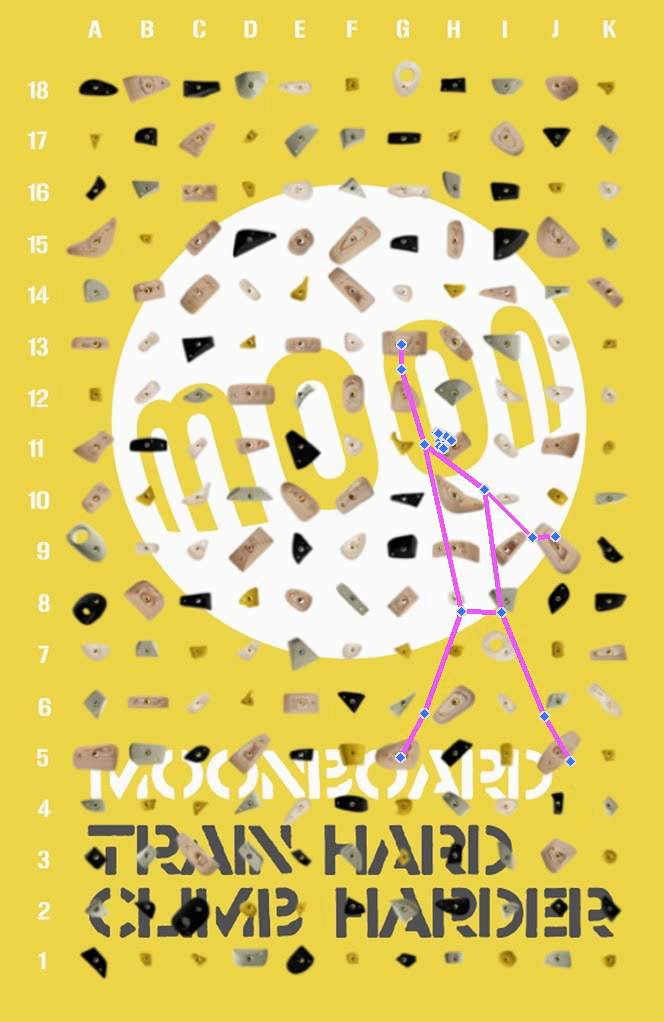}
     \end{subfigure}
     \begin{subfigure}
         \centering
         \includegraphics[width=4cm]{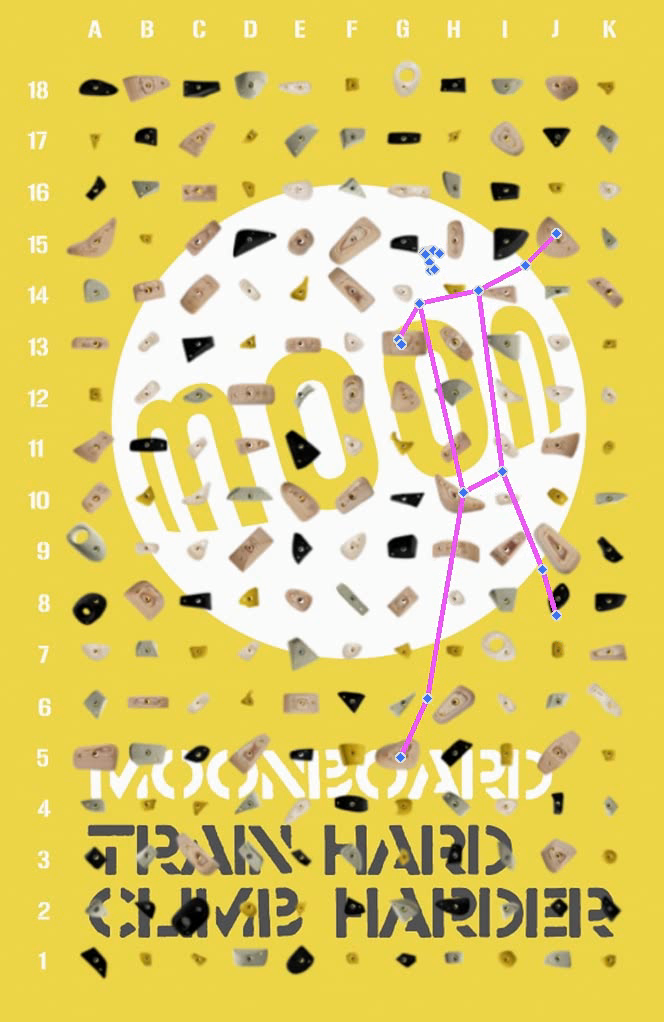}
     \end{subfigure}
     \begin{subfigure}
         \centering
         \includegraphics[width=4cm]{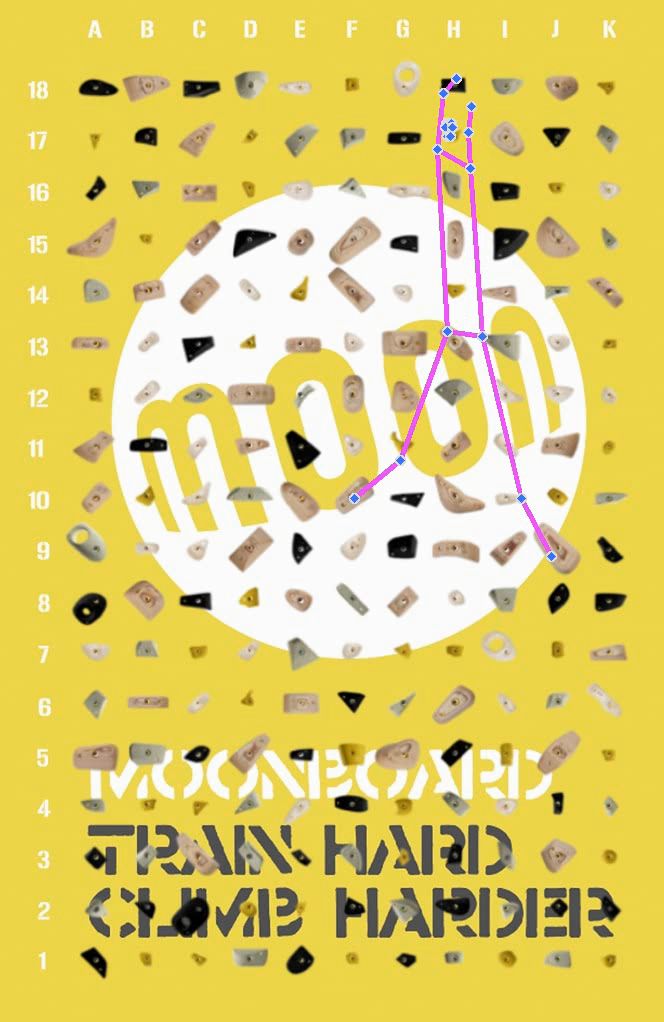}
     \end{subfigure}
  \caption{Generated skeleton from a move sequence, using the procedure described in \autoref{ssec:skeleton_generation}}
\label{fig:skeleton_generated}
\end{figure}

Using \texttt{scikit-learn} \cite{sklearn}'s simple \texttt{Linear Regression} model on a limited number of these videos, we achieve an already very good accuracy of more than $99\%$, enough for our visualization purposes.
From this, we now have access to the whole sequence of coordinates for each landmark, for each frame. Setting all the visibilities  to $1$, we have a complete dataframe of the $99$ landmarks information, for the whole video.
\\

Using \texttt{Mediapipe}, we can now draw the skeleton from the landmarks dataframe and visualize the move sequence. The result is a video of a humanoid figure, climbing the boulder according to the specified move sequence. A few frames of such a video are displayed in \autoref{fig:skeleton_generated}.

The results are definitely satisfactory. We indeed get an idea of how a person would move on the wall, following the given move sequence. However, the use of a linear interpolation implies that the extremities move one after the other, and are fixed when static. Therefore, this can lead to non-physical behaviours, such as limbs stretching, when looking at a dynamic boulder. In general, any problem that would require the climber to lose contact with the wall, or move multiple extremities at the same time, will result in an inaccurate visualization. This is in particular problematic when looking at new-school bouldering, where a lot of jumps and balancing are present. A possible solution to this is an alternative approach, where all the landmarks sequence are directly predicted from the move sequence, instead of relying on linear interpolation for the extremities. This could be the subject of a future work.

\subsection{Selection interface}
\label{ssec:interface}
\begin{figure}[h]
     \centering
     \begin{subfigure}
         \centering
         \includegraphics[width=4cm]{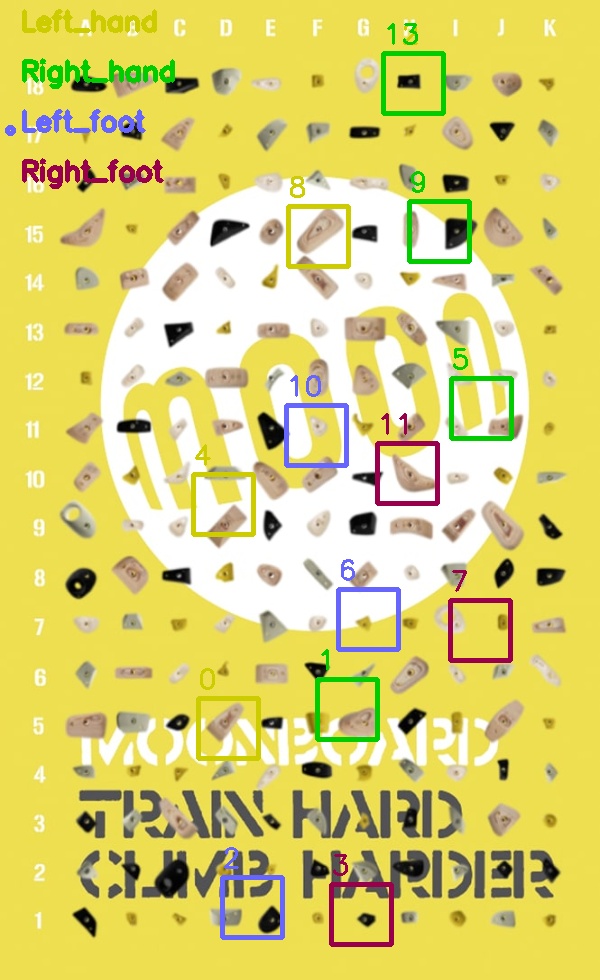}
     \end{subfigure}
     \begin{subfigure}
         \centering
         \includegraphics[width=4cm]{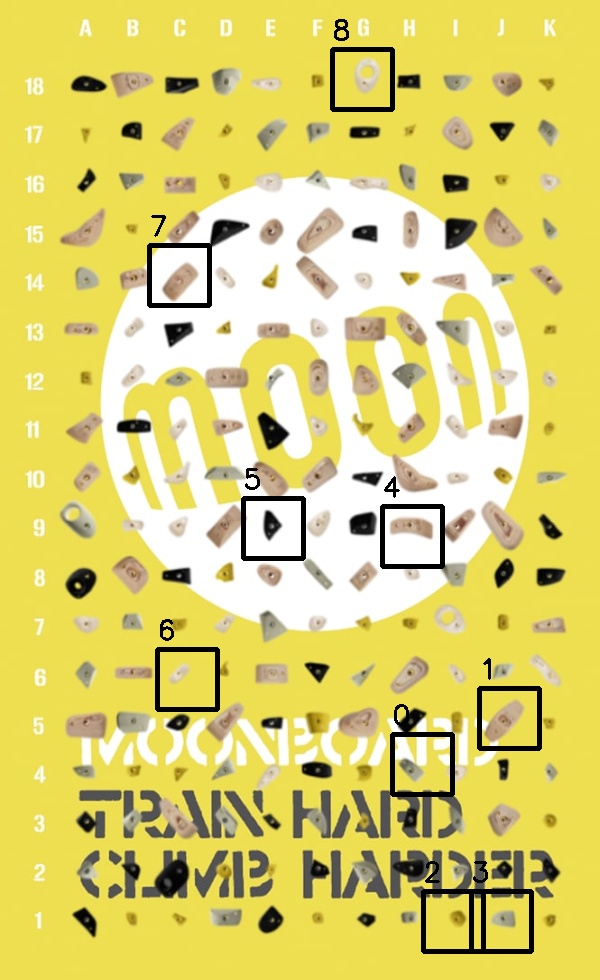}
     \end{subfigure}
  \caption{Selection interface with examples of move sequence (left) and holds sequence (right)}
\label{fig:selection_interface}
\end{figure}

In order to facilitate the use of move sequences for the skeleton pipeline, we developed a selection interface using \texttt{OpenCV}. The goal was to provide a UI allowing the user to dynamically choose their sequence of movements on a standardized wall, the \textit{Moonboard}. By recording the cursor position and the selected extremity, the move sequence can be generated and displayed on the image. Exported as a dataframe, it can then be processed by the above-described pipeline and visualized as a video with the skeleton. 
\\
We report in \autoref{fig:selection_interface} an example of use of the interface. On the left is displayed a selected move sequence, and on the right a simple holds sequence. The difference between the two lies in the fact that the latter is unordered, and does not have extremity information. This has been implemented for future work that will be described in the next section.

\section{Move sequence prediction}
We now move on to the main goal of this project: predicting the move sequence from the holds sequence. This would amount to implementing a systematic solution to bouldering's underlying intellectual challenge of finding the right method (also called \textit{beta}) for a given climb. As we will see, this turned out harder than expected. The general idea was to consider the holds and move sequences as sentences, and use text models to translate one into the other.

\subsection{First try: adaptation of a Sequence to Sequence network}
\label{ssec:seq2seq}
\begin{figure}[h]
    \centering
    \includegraphics[width=6cm]{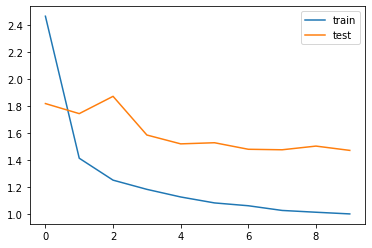}
    \caption{\textit{Seq2seq} model loss trend as a function of the epoch}
    \label{fig:seq2seq_loss}
\end{figure}

We first looked into text translation using sequence to sequence (\textit{seq2seq}) translation models. We refer to \cite{seq2seq} for the basic structure of the network. It is a \texttt{PyTorch} implementation, relying on an \textit{Encoder} and a \textit{Decoder}, combined with an \textit{Attention} mechanism. The data is encoded into a $512$-dimensional latent space, before being processed and mapped onto the output.
\\
The dataset consists of videos used in the students projects, i.e competitions videos provided by the Swiss Olympic Climbing team. The move sequences are extracted from \cite{move_seq_det}, and the holds sequence are computed by clustering the latter with DBSCAN. Regarding our \textit{seq2seq} model, the input sequence is the holds sequence, processed to form a sentence of words of the form "$x_N\_y_N$", while the output sequence is the move sequence, processed in the form of a sentence of words such as "$limb\_x_N\_y_N$".
\\
The model is trained using \textit{Teacher forcing}, and predicts one word at a time from the previous ones. We report in \autoref{fig:seq2seq_loss} the evolution of the loss with the training epochs. The model's performance is regularly evaluated on the validation set with the \textit{Perplexity of Fixed-Length models} metric (PPL):
\begin{equation}
    \mathrm{PPL}({x_0, x_1, \dots, x_t}) = \mathrm{exp}(-\frac{1}{t}\sum_{i}^{t}\mathrm{log} (p_\theta(x_i|x_{<i})))
\end{equation}

\begin{figure}[h]
     \centering
     \begin{subfigure}
         \centering
         \includegraphics[width=4cm]{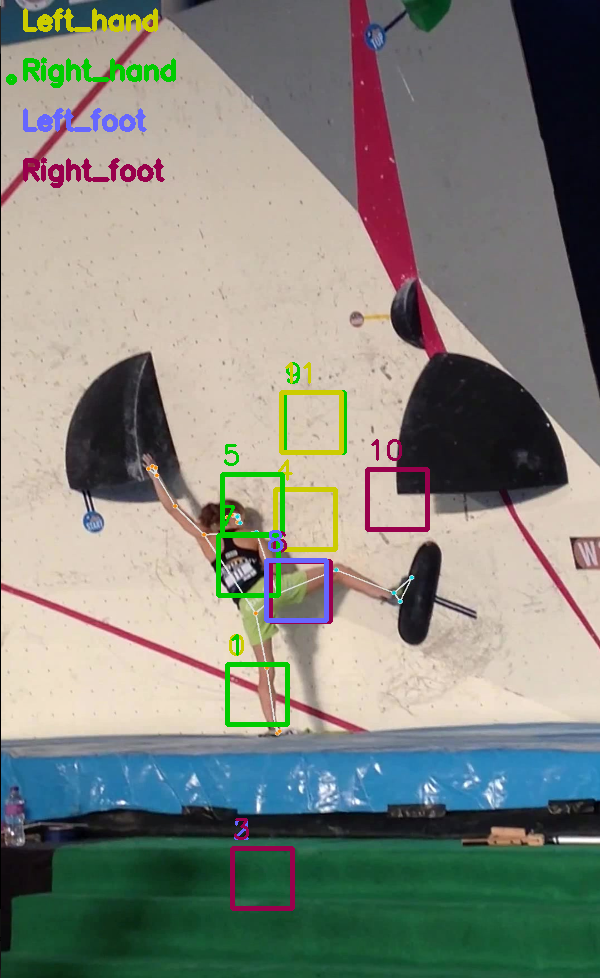}
     \end{subfigure}
     \begin{subfigure}
         \centering
         \includegraphics[width=4cm, height=6.55cm]{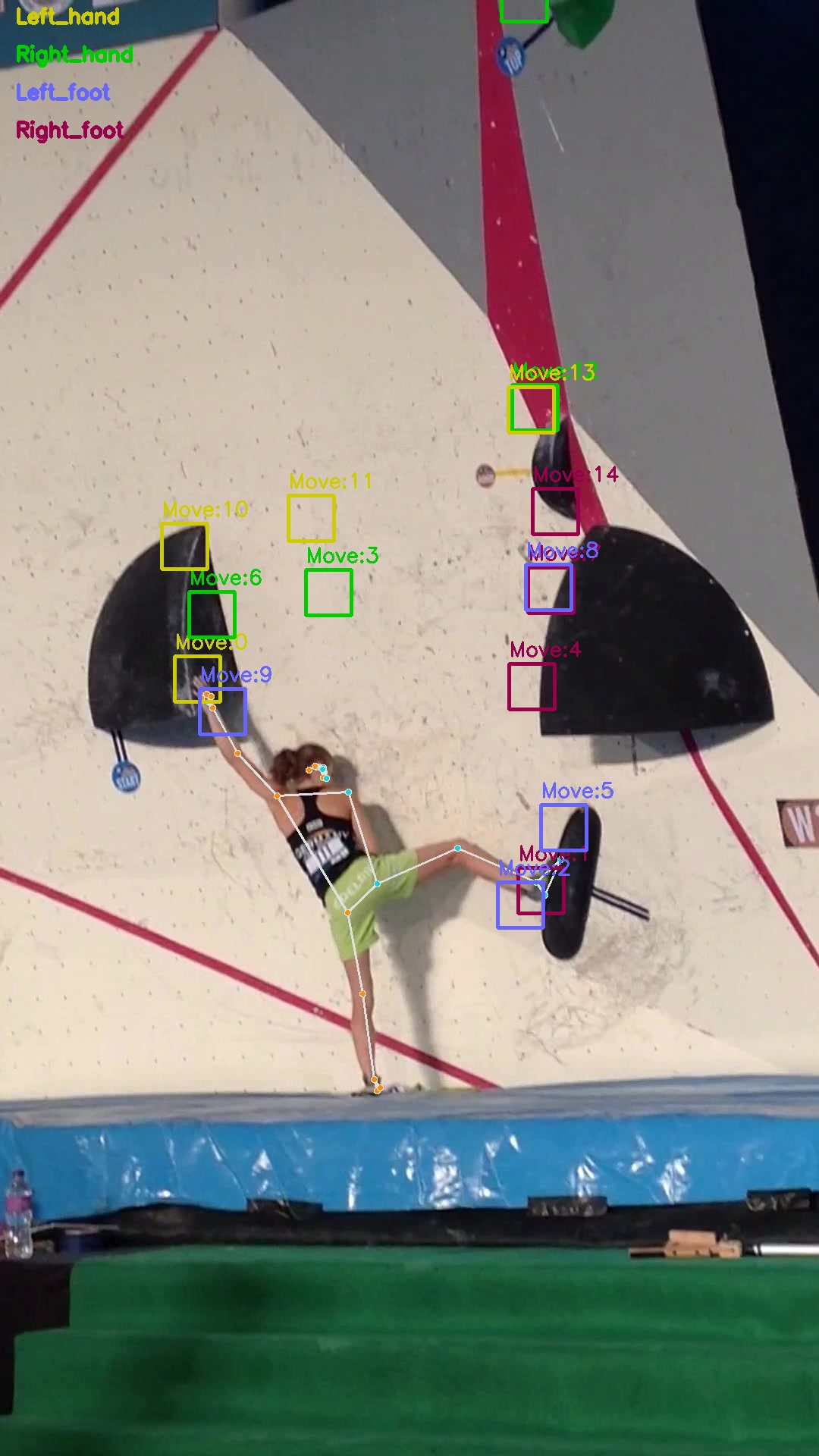}
     \end{subfigure}
  \caption{\textit{Seq2seq} move sequence prediction (left) against the truth (right)}
\label{fig:seq2seq_right_wrong}
\end{figure}

As we can see, the training went well. In fact, in order to really test the model, we trained it for $100$ epochs, which took about two hours on Google Colab's GPU. Nonetheless, the results were disappointing. In fact, as can be seen on \autoref{fig:seq2seq_right_wrong}, the model does not predict the move sequence accurately, and most of the predicted positions are not even holds. This can come from multiple factors, one of them being the technique used to create the holds sequence from the move sequence by clustering. 
\\
Another issue comes from the dataset itself. While the framework designed in \cite{move_seq_det} gives decent results, a lot of move sequences are actually imprecise, with some unpredictability. The videos themselves are not very clean, and exhibit a big diversity of moves. This can impede the model from picking up patterns and accurately learn the translation.
\\
Finally, and arguably the most important, this choice of model constrains us to define the input and output vocabulary to consider every possible combination of \{limb, $x$-coordinate, $y$-coordinate\}. Since the vocabulary is discrete, the coordinates themselves must be discretized on a grid by rounding them up to a given number of decimals. Another constraint is that Colab's memory is limited, so that we cannot create arbitrarily big vocabularies. In fact, we were only able to make it work by rounding the coordinates to \textbf{one decimal}, i.e. a coordinate can only take the values $\{0, 0.1, 0.2, ..., 1\}$. This hugely impacts the performance of the model, since it can only predict moves on this grid, which can be far from the actual position. This can in particular explain why the prediction in \autoref{fig:seq2seq_right_wrong} is so different from the truth, the predicted positions not being accurate.
\\
It is worth mentioning that using the visualization pipeline described in \autoref{ssec:skeleton_generation}, the movements of the generated skeleton are not that far off the real climber's, keeping in mind that the holds are not correctly placed.

\subsection{Next try: Autoregressive Transformer model with Positional Embedding}
Since the previous model did not give satisfying results, we tried looking at a simplified problem: removing the extremity information, and just focusing on the holds ordering. In this context, the input is a sequence of holds represented by their coordinates; and the output is the same sequence, but sorted to follow the order in which these holds are used by the climber.
\\
In order to do so, we considered a \textit{Transformer} model, which also involves an Encoder-Decoder structure, but the decoder is a simple linear layer. The main difference between this and the model described in \autoref{ssec:seq2seq} is that this is a sequence completion model, instead of a sequence translation one. 
\\
\subsubsection{Data formatting}
The data used with this model is different than the one used before. As previously mentioned, the competition videos were not easy to work with. They are very different from one to the other, and the move sequence detection is not fully reliable. Moreover, extracting the holds information is not fully precise either. Therefore, in the following, we consider a different dataset. It consists of $20$ videos of a climber on the Moonboard, taken from \cite{playlist}. The holds sequences have been manually extracted using the interface presented in \autoref{ssec:interface}. Since the model should be able to sort the holds, the output should not depend on the order of the input. Therefore, we added random permutations of the input sequences. This ensures invariance to the holds sequence order, and extends the dataset. Using $50$ permutations allows to extend the dataset to $1000$ sequences.
\\
This model handles the data in a different way with respect to the other one. More precisely, the input and output sequences are actually concatenations of the original and sorted holds sequences. The input is made out of all the original elements, and all the sorted elements except the last one. The output is made out of all the original elements starting from the second, and all the sorted elements. We report in \autoref{tab:data_transformer} an example of such data. For clarity, we only report the coordinates, but note that the Transformer actually works with \textbf{tokens} ($i.e.$ words). In our case, since we aim at predicting the order of the holds, the input tokens are simply the numbers $[0,N-1]$ and the output tokens are the shuffled versions of those according to the output sequence order. The coordinates are used in another way, which will be described in the next section.

\begin{table}[!htb]
    \begin{minipage}{.23\linewidth}
      \caption*{Original sequence}
      \centering
        \begin{tabular}{|c|c|c|}
            \hline
            $x_1$ & $y_1$ \\
            \hline 
            $x_2$ & $y_2$ \\
            \hline 
            $x_3$ & $y_3$ \\
            \hline 
            $x_4$ & $y_4$ \\
            \hline 
            $x_5$ & $y_5$ \\
            \hline
        \end{tabular}
        $\bigoplus$
    \end{minipage}
    \begin{minipage}{.23\linewidth}
      \centering
        \caption*{Ordered sequence}
        \begin{tabular}{|c|c|c|}
            \hline
            $x_4$ & $y_4$ \\
            \hline 
            $x_2$ & $y_2$ \\
            \hline 
            $x_5$ & $y_5$ \\
            \hline 
            $x_3$ & $y_3$ \\
            \hline 
            $x_1$ & $y_1$ \\
            \hline
        \end{tabular}
    $\xrightarrow{}$
    \end{minipage} 
    \begin{minipage}{.23\linewidth}
      \centering
        \caption*{Input sequence}
        \begin{tabular}{|c|c|c|}
            \hline
            $x_1$ & $y_1$ \\
            \hline 
            $x_2$ & $y_2$ \\
            \hline 
            $x_3$ & $y_3$ \\
            \hline 
            $x_4$ & $y_4$ \\
            \hline 
            $x_5$ & $y_5$ \\
            \hline
            $x_4$ & $y_4$ \\
            \hline 
            $x_2$ & $y_2$ \\
            \hline 
            $x_5$ & $y_5$ \\
            \hline 
            $x_3$ & $y_3$ \\
            \hline
        \end{tabular}
        ,
    \end{minipage} 
    \begin{minipage}{.23\linewidth}
      \centering
        \caption*{Output sequence}
        \begin{tabular}{|c|c|c|}
            \hline 
            $x_2$ & $y_2$ \\
            \hline 
            $x_3$ & $y_3$ \\
            \hline 
            $x_4$ & $y_4$ \\
            \hline 
            $x_5$ & $y_5$ \\
            \hline
            $x_4$ & $y_4$ \\
            \hline 
            $x_2$ & $y_2$ \\
            \hline 
            $x_5$ & $y_5$ \\
            \hline 
            $x_3$ & $y_3$ \\
            \hline
            $x_1$ & $y_1$ \\
            \hline
        \end{tabular}
    \end{minipage} 
    \caption{Transformer data}
    \label{tab:data_transformer}
\end{table}

Not all sequences have the same length. To allow the model to work with similar-sized tensors, a padding has been applied, where we added an imaginary hold. We also tried padding the sequences with the last holds in the input but the results were similar.
\subsubsection{Model specifics}
This Transformer works thanks to two main components:
\begin{enumerate}
    \item An \textit{Attention Mask}: this object allows to specify which part of the input should be used at inference. In particular, the model will try to predict one token at a time, using the previous ones. Therefore, it will first try to predict the original tokens, then will arrive at the first output token, and give a prediction for it based on the already predicted ones. The Attention Mask is used to tell the model that it should only look at the tokens before the current position to make the prediction. When evaluating the model on completely new datapoints, we just have the original sequence, so the part related to the sorted one in the input and output sequence can be replaced by random values, since they are anyway updated at inference each time the model makes a prediction.
    \item A \textit{Positional Embedding}: in common text processing, this object allows the model to take context, $i.e.$ the word order, into account when making its prediction. In our case, we do not really care about the tokens (numbers from $0$ to $N-1$) because they do not hold any information regarding the output. What matters most are actually the \textbf{coordinates} of the holds. Therefore, we adapted the usual Positional Embedding described in \cite{posemb} to work with the coordinates instead, adding their representation to the token embedding.
\end{enumerate}

\begin{figure}[h]
    \centering
    \includegraphics[width=8cm]{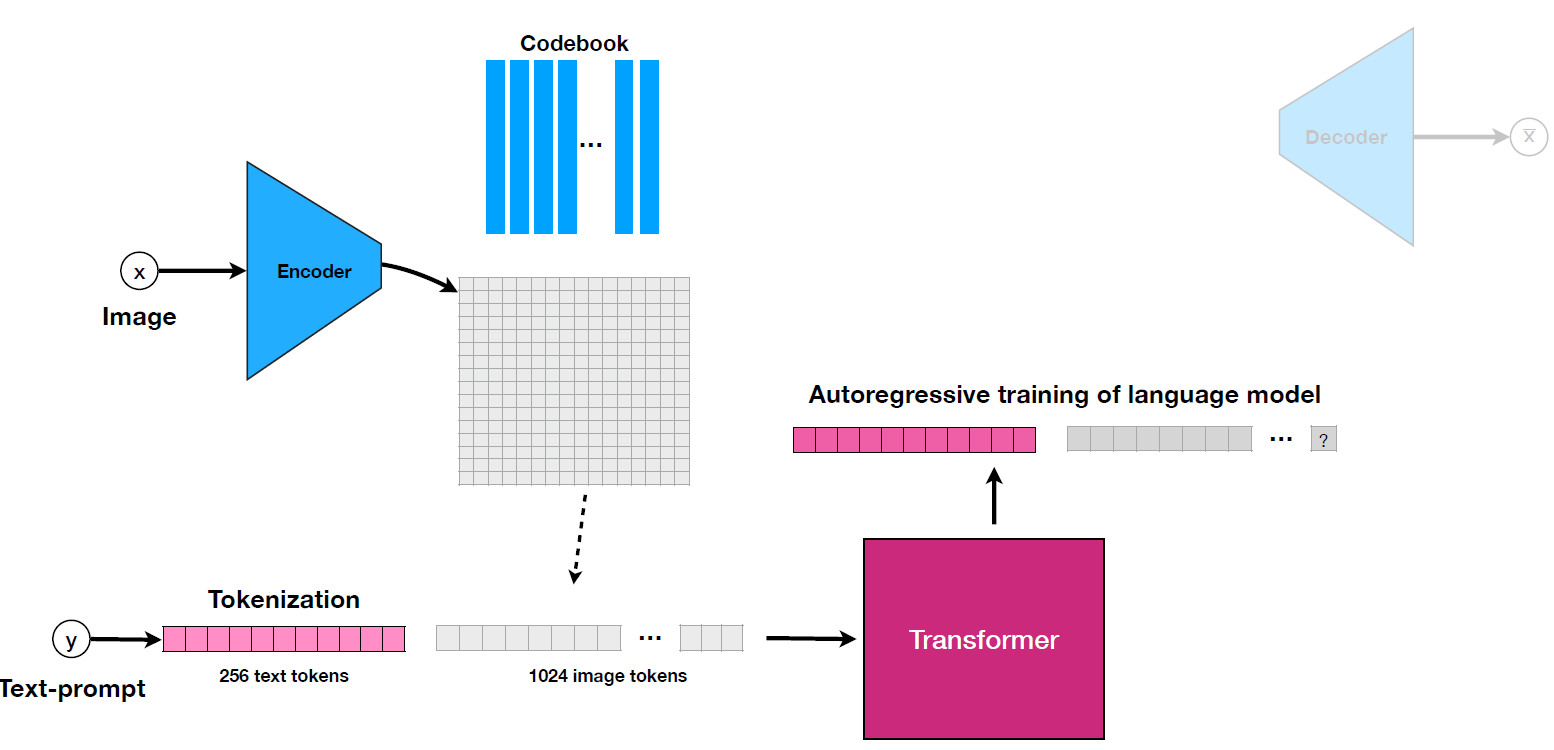}
    \caption{Autoregressive Transformer functioning \cite{dall}}
    \label{fig:transformer_diag}
\end{figure}

\autoref{fig:transformer_diag} reports the general functioning of an Autoregressive Transformer for sequence completion, in the context of text processing.
\subsubsection{Training and results}
The model has been trained with the \textit{Cross Entropy} loss and the \textit{Adam} optimizer. Experimenting with various learning rate, embedding dimensions, and Positional Embeddings, the obtained results were disappointing. Most notably, the model almost always outputs the padding token no matter which input is fed into it. During training, both validation and training losses go down, but quickly stabilize around a fixed value. Since most sequences are shorter than the maximum length, the padding element is repeated a high number of times, and the model can obtain a consistent average accuracy by repeatedly predicting it.
\subsection{Final try: Simplified Transformer with direct Forward Pass}
The last model we tried is a simpler version of the previous one. No Positional Embedding is used, and the model does not try to complete the sequence one token at a time using the mask. Instead, the model consists of a succession of a encoding layer followed by a decoder linear layer, and directly outputs a prediction in a single forward pass. The data is therefore made out of the original and sorted sequences, without the need to concatenate them. 
\\

\begin{figure}
    \centering
    \includegraphics[width=8cm]{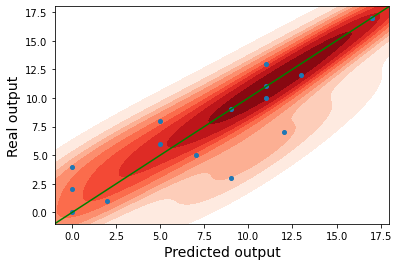}
    \caption{Evaluation of the model's performances on a validation sequence of size $14$, padded to a fixed length of $17$. Accuracy $\approx 35\%$.}
    \label{fig:kde}
\end{figure}

We trained the model with \textit{Adam} for $300$ epochs, experimenting with different learning rates. The best model (evaluated by the validation loss), was saved at regular intervals and then evaluated on the validation set. The results were here very different from the previous ones, and more satisfying. In fact, the model does not always predict the padding element ($17$), but actually outputs token numbers for the existing holds. However the results look pretty random, with only very few of them being accurate. We report in \autoref{fig:kde} an example of an output/target combination on a given sequence. While we do witness a trend, we remind that the results are \textbf{discrete tokens} and not continuous values. The model's accuracy on this sequence is about $35\%$, but this is mainly due to the fact that the last $3$ tokens are the padding one, and the model was able to capture them. On the other hand, only $2$ non-padding tokens have been accurately predicted. Nonetheless, this model is an improvement on the previous ones, and more investigation on it could possibly lead to better predictions for this problem.

\section{Conclusions}
In this work, we gave a first insight into the application of Machine Learning techniques to bouldering, experimenting with move sequence visualization and prediction. We developed an interactive pipeline for visualization, and investigated different models for holds sequence sorting, a first step towards move sequence prediction. As of now, the results are not satisfying and would require more research in the future. Nevertheless, they represent a first step into this kind of approach, and can be extended as part of a future work.

\bibliographystyle{IEEEtran}
\bibliography{bibliography.bib}
\end{document}